\pdfoutput=1
\documentclass[11pt]{article}

\usepackage[]{acl} 

\usepackage{times}
\usepackage{latexsym}
\usepackage{amsmath}
\usepackage{stfloats}

\usepackage[T1]{fontenc}
\usepackage[utf8]{inputenc}

\usepackage{microtype}

\usepackage{color, colortbl}
\usepackage[colorlinks]{}
\usepackage{enumitem}
\usepackage{graphicx}
\definecolor{light-gray}{gray}{0.9}
\title{Controllable Generation of Dialogue Acts for Dialogue Systems via Few-Shot Response Generation and Ranking}

\author{Angela Ramirez \and Kartik Aggarwal \and Juraj Juraska \\ \and \bf Utkarsh Garg \and \bf Marilyn A. Walker \\ 
  University of California Santa Cruz \\ \texttt{aramir62, kartik, mawalker@ucsc.edu} 
  \\}
\begin{document}
\maketitle
\begin{abstract}
Dialogue systems need to produce responses that realize multiple types of dialogue acts (DAs) with high semantic fidelity. In the past, natural language generators (NLGs) for dialogue were trained on large parallel corpora that map from  a domain-specific  DA and  its semantic attributes to an output utterance. 
Recent work shows that  pretrained language models (LLMs) offer new possibilities for controllable NLG using prompt-based learning.  
%Previous work on semantically-controlled NLG has explored many    methods for improving the semantic accuracy of  NLGs for dialogue. However, in conversation, the same set of  semantic attributes can be realized by different DAs. 
Here we develop a novel few-shot overgenerate-and-rank  approach that achieves the controlled generation of DAs.  We compare eight few-shot prompt styles that include a novel method of generating from textual pseudo-references  using a textual style transfer approach. %,  a second novel approach that provides definitions of DAs in the prompts, and a baseline that  linearizes the DA representation.   
We develop six automatic ranking functions that identify outputs with both the correct DA and high semantic accuracy at generation time. We test our approach on three domains and four LLMs. To our knowledge, this is the first work on NLG for dialogue that automatically ranks outputs using both DA and attribute accuracy. For completeness,  we compare our results to fine-tuned few-shot models trained with 5 to 100 instances per DA.  Our results show that several prompt settings achieve perfect DA accuracy, and near perfect semantic accuracy (99.81\%) and perform better than few-shot fine-tuning.
%(97.7\% semantic accuracy and  80.6\% DA accuracy). 
\end{abstract}

\section{Introduction}

Dialogue systems need to faithfully produce utterances that realize multiple types of dialogue acts (DAs), such as providing opinions, making recommendations, or requesting information. In the past, natural language generators (NLGs) for dialogue have   been trained  on large parallel corpora that map from  a domain-specific meaning representation (MR) that specifies the desired DA and  semantic attributes to an output utterance. The  NLG must faithfully generate utterances that realize the style and form of the DA, and all of the specified attributes, as shown by the reference utterances in Table~\ref{tab:ex_video_game_dataset-3DAs-sameMR}. 
Recent work shows  that  pretrained language models (LLMs) offer new possibilities for controllable NLG using prompt-based  learning (PBL) \cite{brown2020language,radford2019language,liu_surveyprompts}.
Here we present a novel few-shot overgenerate-and-rank  approach that achieves the controlled generation of DAs. 

\begin{table}[htb]
\def\arraystretch{1.1}
\begin{small}
\begin{center}
\begin{tabular}{p{3.0in}}
    	\hline
    	\rowcolor{light-gray}   {\bf \emph{Attributes and Values}}  \\
    	(\textsc{name} [\textbf{Call of Duty: Advanced Warfare}], \textsc{rating} [\textbf{excellent}], \textsc{developer} [\textbf{Sledgehammer Games}], \textsc{esrb} [\textbf{M (for Mature)}]) 
     \\ 
   %  \hline 
    	% I think that  \textbf{Call of Duty: Advanced Warfare} from SledgeHammer games is \textbf{one of the best the M rated games} I've ever played. It's really addictive. 
      	\rowcolor{light-gray} {\bf   \emph{give\_opinion}}  \\
       \textbf{Call of Duty: Advanced Warfare} must be \textbf{one of the best games} I've ever played. \textbf{Sledgehammer Games} always nail their \textbf{M-rated} games. \\ 
       \hline
    	 \rowcolor{light-gray}  {\bf   \emph{recommend}} \\
    	%\emph{recommend}(\textsc{name} [\textbf{Call of Duty: Advanced Warfare}], \textsc{rating} [\textbf{excellent}], \textsc{developer} [\textbf{Sledgehammer Games}], \textsc{esrb} [\textbf{M (for Mature)}]) \\
        %\hline
    	% Speaking of \textbf{M rated} games developed by \textbf{Sledgehammer Games}, have you tried \textbf{Call of Duty: Advanced Warfare}? \\
        Since you seem to \textbf{love} \textbf{M-rated} games developed by \textbf{Sledgehammer Games}, I wonder if you have tried \textbf{Call of Duty: Advanced Warfare}. 
        \\ \hline
      \rowcolor{light-gray}  {\bf   \emph{inform}} \\
        %make an inform that uses the same attributes as the recommend to make our point.
   	   % \emph{inform} (\textsc{name} [\textbf{Call of Duty: Advanced Warfare}], \textsc{rating} [\textbf{excellent}], \textsc{developer} [\textbf{Sledgehammer Games}], \textsc{esrb} [\textbf{M (for Mature)}]) \\
       % \hline
    	% Developed by \textbf{Sledgehammer Games}, \textbf{Call of Duty: Advanced Warfare} is targeted at \textbf{mature} audiences.  \\
        Developed by \textbf{Sledgehammer Games}, \textbf{Call of Duty: Advanced Warfare} is targeted at \textbf{mature audiences} and has overall \textbf{very positive ratings}. \\
 \hline
    \end{tabular}
     \end{center}
    \end{small}
	\caption{Sample ViGGO dialogue acts (DAs)  \cite{juraska2019viggo}. The same attributes and values can be realized as different DAs.   
 %The DA is in italics,  the attribute names are in small caps, and their  values  are bolded. The bolded attributes in the reference utterances  illustrate their faithful realization.
 }
    \label{tab:ex_video_game_dataset-3DAs-sameMR}
   
\end{table}

%juraska-walker-2021-attention
Previous work on semantically-controlled NLG has focused on  improving semantic accuracy 
\cite{rastogi2020towards,xu2021augnlg,du2022self,wen2015semantically,kedzie2020controllable,juraska-walker-2021-attention}. 
%%{\it inter alia}. 
However, 
Table~\ref{tab:ex_video_game_dataset-3DAs-sameMR}  shows how the  the same set of semantic attributes can be realized by different DAs, such as {\it give\_opinion}, {\it recommend} and {\it inform}, each of which affect the dialogue state differently  \cite{traum1994discourse}.
%name, rating, developer and ESRB  semantic attributes, but each DA updates the dialogue state differently, and  leads to different user response types. 

Obviously an NLG for  dialogue  needs to faithfully realize the DA as well as the semantic attributes.  
%Here we present a novel few-shot overgenerate-and-rank NLG engine that can automatically identify at inference time which outputs have both the correct DA and high semantic attribute accuracy.  
However,  previous work  has neither  {\it controlled for}  nor {\it evaluated} DA accuracy.  We speculate that this is because many NLG training sets, such as  E2E, Weather, WebNLG, {Wiki\-Bio}, DART and ToTTo, 
%are not intended for dialogue systems, and thus 
only include {\it inform}  DAs~\cite{novikova-etal-2017-e2e,belz2008automatic,gardent-etal-2017-creating,lebret2016neural,nan-etal-2021-dart,parikh2020totto}.  
%Because there is only one type of DA, the MRs for such datasets don't specify any DA  at all. 
Yet NLG training sets  for spoken dialogue include many  types of DAs,  e.g. the ViGGO corpus has 9   DAs \cite{juraska2019viggo},  the RNNLG corpus provides 13   DAs  \cite{wen2015semantically},  
MultiWOZ has 34   DAs \cite{eric-etal-2021-multi}, and  Topical Chat was automatically labelled with 11   DAs 
 \cite{hedayatnia2020policy,mezza2018iso}.

%({\color{red} also in the context of PBL, what previous work has used ranking for SAC?}) 

We present a few-shot  PBL framework  that overgenerates and ranks NLG outputs and achieves high accuracy for both semantic attributes and DAs.   We develop high accuracy DA classifiers for three   domains and use them to define 6 ranking functions that combine estimates of   DA probability with measures of  semantic accuracy.
We also  compare a combination of prompt formats,  prompt sampling methods, and DA representations. Several prompt templates take the novel approach of treating DA control as a textual style transfer (TST)  problem \cite{reif-etal-2022-recipe}. 
%We vary the DA representation within the prompt by simply linearizing the MR  \cite{wen2015semantically,harrison2019maximizing} or by generating textual pseudo-references.  We vary prompt formats to simply use one of the representations of the MR, and compare this with several types of prompts used  for textual style transfer (TST)  \cite{reif-etal-2022-recipe}. Our TST prompts treat DA types as different styles to be controlled. We also create  novel prompts that provide definitions of DAs 
For completeness,  we report results for few-shot fine-tuned  models trained with  5 to 100 instances per DA.  Our contributions include: 
\begin{itemize}
\item The first results  showing that dialogue acts can be controlled with PBL;
\vspace{-.12in}
\item A new overgenerate-and-rank framework that automatically ranks generation outputs  for DA accuracy at generation time;
\vspace{-.1in}
\item A systematic exploration of both domain-specific and general measures in ranking functions, and a comparison of their performance;
\vspace{-.3in}
\item Results showing that a ranking function that prioritizes  DA  correctness results in higher semantic accuracy. 
\vspace{-.1in}
\item The definition of novel textual DA representations that support automatic ranking for semantic accuracy using off-the-shelf metrics such as BLEU and Beyond-BLEU; 
\vspace{-.1in}
\item The systematic testing of 8  prompt formats that re-cast data-to-text generation as a text-to-text task, and an examination of their performance across 4  LLMs.
\end{itemize}

The results demonstrate large performance differences across prompt styles, but show that many prompts achieve perfect DA accuracy, and  semantic accuracy as high as 99.81\% with only 10 examples, while 100-shot per DA fine-tuning only achieves 97.7\% semantic accuracy, and 80.6\% DA accuracy. 
%Our  results also show that formulating the data-to-text task as textual style transfer using pseudo-references yields the highest performance. 

\section{Related Work}
\label{rel-work-sec}

This paper applies few-shot PBL to the task of controllable generation of DAs using an overgenerate-and-rank NLG framework. The  overgenerate-and-rank paradigm for NLG   has primarily used  two  methods for ranking: (1) language model probability \cite{langkilde1998generation}; and (2) ranking functions trained from human feedback \cite{rambow2001evaluating,bangalore2000evaluation,liu2016not}. We extend this framework by applying it in the context of PBL, by using DA probability in  ranking, and by comparing many ranking functions, including Beyond-BLEU and BLEU baselines \cite{wieting-etal-2019-beyond,papineni2002bleu}. 

We know of only a few previous studies on controllable generation of  DAs in the context of dialogue systems, each of which has only
focused on one or two types of DAs.
Obviously, tasks like question generation (QG)
aim at controllable generation of questions \cite{harrison2018neural,zhang2021review} but research on QG is not focused
on trying to control
the generation of questions as opposed to other types of DAs.  However, some work has focused on controlling questions in dialogue, e.g. 
\citet{hazarika2021zero} learned a latent representation of questions from a labelled corpus and then used this as a prompt prefix to control 
question generation. \citet{see2019makes} fine-tuned a Persona Chat model and  tested decoding
methods that controlled question frequency, but did not guarantee a question on a particular turn.  Other work has focused on dialogue acts like opinions and recommendations. For example, \citet{oraby2019curate} curated opinionated utterances from user
reviews that had been marked with exclamation points, and then used the exclamation points as a way to control the production of exaggerated opinions. 
\citet{reedetal20} used token supervision to control the production of {\tt recommendation} as opposed to {\tt inform} dialogue acts where {\tt recommendation} DAs stated that a particular restaurant was the best and then justified the recommendation with attributes from the MR. \citet{ramirez2023controlling} used PBL with similar prompts to control the expression of Big 5 personality types \cite{harrison2019maximizing}, rather than dialogue acts. 

It is well known that data-to-text NLGs based on fine-tuned LLMs are prone to semantic errors  \cite{ji-hallucination,rashkin-etal-2021-increasing}, thus previous work has focused on  methods for ensuring semantic correctness.
This includes  automatically augmenting the training data \cite{xu2021augnlg,du2022self},  modifying the input representation \cite{kedzie2020controllable,heidari-etal-2021-getting}, using rankers or classifiers or decoding methods that identify semantically accurate or acceptable candidates \cite{harkous-etal-2020-text,juraska-walker-2021-attention,wen2015semantically, shen-etal-2019-pragmatically,batra-etal-2021-building}. 
Previous work on few-shot PBL for semantically-controlled NLG has not attempted to 
control DA  accuracy \cite{reed2022jurassic,soltan2022alexatm}, and has not used an overgenerate and rank approach, resulting in lower semantic accuracies than we report here.

Much  previous work on few-shot NLG has investigated few-shot finetuning rather than few-shot PBL. 
Previous work on the ViGGo, TV and Laptop corpora  \cite{xu2021augnlg,du2022self,kedzie2020controllable,juraska-walker-2021-attention} supports  direct comparison to our work, but is not few-shot, does not rank outputs or use PBL. 
FewShotWoz 
trains a model called SC-GPT on a 400K data-to-text corpus, and then tests transfer learning
with only 40 or 50  fine-tuning examples \cite{peng-etal-2020-shot}. Other recent work develops methods for augmenting FewShotWoz using synthetic data or by self-training and shows improvements in semantic accuracy and BLEU score. The FewShotWoz corpus includes many   types of DAs 
but none of this previous work includes an evaluation of NLG DA accuracy. 
Previous work on few-shot finetuning  in the weather  domain used 300 examples in fine tuning, and also explored different ways of  textualizing the MR   \cite{heidari-etal-2021-getting}, but  did not attempt to control DAs, develop ranking functions, evaluate DA accuracy, or use  instructions such as  our novel definitional prompts and the templates for TST tasks. \citet{heidari-etal-2021-getting} achieve an 85\% reconstruction accuracy, while our best prompt/LLM combinations achieve
99.44\% PERF score for ViGGO, 99.57\% PERF for TV and 99.47\% PERF for Laptop, a similar metric to reconstruction accuracy, with only 10 examples. 
%Previous work on few-shot data-to-text on WikiBio has used between 50 and 500 training examples for fine-tuning, with only the {\it inform} DA    \cite{chen-etal-2020-shot}. 

%\vspace{-.1in}
\section{Automatically Ranking NLG Outputs}
\label{sec-auto-ranking}

We start by providing a mathematical formulation of our problem. When generating from a DA representation, 
a high-quality response should: (1) manifest the specified DA; (2) have no missing or incorrect mentions of the attributes; (3) hallucinate no additional attributes; and (4) be fluent. Thus the generated utterance $y$, conditioned on an input $x$ composed of DA $d$ and attribute values $a$, can be formulated as $y = f(d,a)$. The conditional likelihood of $y$  given the MR can then be decomposed using Bayes Rule into the product of three probabilities: 
%{\begin{align*}  
% \label{likelihood}
% p(y | x)\\
% & = p( y | d,s) \\
%& = \frac{p(y, d, s)}{p(d, s)} \, \propto p(y, d, s)\\
%  & = p(d|y,s) * p ( y,s) \\
%  & = p(d|y,s) * p (s| y) * p(y)
 %\end{align*}}
{\begin{equation}  
 \label{likelihood}
 p( y | d,a)  = p(d|y,a) * p (a| y) * p(y)
 \end{equation}
 }

The term $p(d|y,a)$ is the DA probability given the generated utterance $y$ and the semantic attributes $a$. The  term $p(a|y)$ represents the semantic accuracy. %, which can be measured in many   ways. 
The term $p(y)$ is the unconditional probability of the generated utterance, which is commonly used as a measure of fluency. Below, we show how we compute  estimates of these terms at generation time, and then explain their use in the ranking functions.
%We calculate the DA probability with a DA classifier (Sec.~\ref{DA-tagger-sec}), compare multiple ways of  computing semantic accuracy   (Sec.~\ref{SACC-sec}), and compute fluency as  sentence probability (Sec.~\ref{ranking-funcs}).

%\subsection{Dialogue Act Classifier}
\label{DA-tagger-sec}
\vspace{0.1in}
\noindent{\bf Dialogue Act Classifier.} 
The term $p(d|y,a)$ requires highly accurate DA classifiers to use in automatic ranking. 
We fine-tuned two classifiers using pre-trained bert-base-uncased on HuggingFace. We discovered that even though the  ViGGO, Laptop and TV training corpora are good size \cite{juraska2019viggo,wen2015semantically}, producing high accuracy classifiers
required us to modify the training data.\footnote{We also experimented with training classifiers
for MultiWoz but were unable to get high accuracies  due to noise in DA labelling,
which is known to be an issue with MultiWoz \cite{zou-2022-multi}.} We originally trained the ViGGO classifer with
the original ViGGO training set,  when we applied this  classifier to the generated outputs, we noticed many cases of low confidence classification. A qualitative analysis of the data showed that  many generated outputs did not actually fit into the original ViGGO ontology, which is not surprising, given that the training data for an LLM would have included many different types of DAs. 
%Appendix Section~\ref{appendix-incorrect-das-sec} provides examples of generated DAs outside the ontology and incorrectly generated DAs.

To increase the ViGGO classifier performance, we introduced an "Other" class of dialogue acts,  doubly annotated another 1000 ViGGO NLG outputs by hand, and added them to the original training set. Final results are shown in Table~\ref{tab:ViGGO-DAClassifierResult}.

\begin{table}[htb]
%\begin{wraptable}{r}{2.0in}
\centering
%\scalebox{0.6}{
\def\arraystretch{1.1}
\small
\begin{tabular}{|r|r|r|r|}
\hline
% \multicolumn{1}{c|}{} & \multicolumn{4}{c|}{\textbf{Train/Test}\\
\textbf{Dialogue Act} & \bf ViGGO  \\\hline
{\it confirm} & 0.99   \\
{\it inform} &  0.98 \\
{\it suggest} &   0.91 \\
{\it give\_opinion} & 0.90 \\
{\it recommend} &  0.92\\
{\it request} &   0.94 \\
{\it request\_attribute} &  0.93 \\
{\it request\_explanation} & 0.99 \\
{\it verify\_attribute} & 0.94 \\
{\it other} &   0.78\\\hline \hline
Weighted Average &  0.97 \\ \hline
\end{tabular}%
%}
\caption{ViGGO DA classification F1 scores.}
\vspace{-0.2in}
\label{tab:ViGGO-DAClassifierResult}
\end{table}

The second classifier was trained using the complete RNNLG corpus with all 4 domains to maximize classifier domain transfer. When we tested it on the RNNLG test set, we discovered that several classes had low F1. Examination of the confusion matrix showed that the {\it recommend} and {\it inform} DAs were highly confusable, so we created a new type of DA we call ``describe'' by combining their training sets. The final results for 
for the RNNLG classifiers is shown in 
Table ~\ref{tab:RNNLG-DAClassifierResult}. 
%column 1 (OV-OV) shows the results:  the lowest F1 is  
%.91  for the {\it request} DA, for an average F1 of .98. 

\begin{table}[htb]
\centering
%\scalebox{0.6}{
\centering\def\arraystretch{1.1}
\small
\begin{tabular}{|r|r|r|r|}
\hline
% \multicolumn{1}{c|}{} & \multicolumn{4}{c|}{\textbf{Train/Test}\\
\textbf{Dialogue Act} & \bf Laptop & \bf TV   \\\hline
{\it compare} & 1.00 & 1.00 \\
{\it confirm} & 0.96  &  0.95 \\
{\it describe} & 1.00  &  1.00 \\
{\it inform all} &0.86  &  0.92 \\
{\it inform count} & 1.00  & 1.00 \\
{\it inform no info} & 1.00  & 1.00 \\
{\it inform no match} & 0.98  &  0.94\\
{\it inform only match} & 0.83  & 0.87 \\
{\it suggest} & 1.00  &  1.00\\\hline \hline
 Weighted Average &  0.99 & 0.99\\ \hline
\end{tabular}%
%}
\caption{Laptop and TV DA classification F1 scores. The {\it describe} DA $=$ combination of the {\it inform} and {\it recommend} DAs in  the original dataset.}
\vspace{-0.1in}
\label{tab:RNNLG-DAClassifierResult}
\end{table}

We provide these DA classifiers along with additional human-labelled  model outputs  so that other researchers can  duplicate our setup.\footnote{\scriptsize\url{https://github.com/aramir62/da-nlg}} The resulting classifiers achieve average F1s over .97 for all three domains.

%\subsection{Semantic Accuracy}
\vspace{0.1in}
\label{SACC-sec}
\noindent{\bf Semantic Accuracy.} Work on data-to-text NLG  often computes semantic accuracy as the Slot Error Rate (SER), i.e., the percentage of slots across all outputs $y$ that the NLG realized incorrectly, with models either carefully tuned by hand, or trained by artificially creating incorrect realizations \cite{wen2015semantically,dusek2019semantic,juraska2018deep,reedetal20,wiseman2017challenges,harkous-etal-2020-text,kedzie2019good,kedzie2020controllable}.
There is a toolkit for  SER  for all three domains,\footnote{\scriptsize\url{https://github.com/jjuraska/data2text-nlg}} which we use to calculate  SACC:
  \begin{equation}
     \text{SACC} = 1 - \text{SER}
 \end{equation}

Because the SACC scripts are domain specific, we also create new metrics that are based on  BLEU, BLEURT, Beyond-BLEU and BertScore, widely used  measures of semantic accuracy and semantic preservation \cite{papineni2002bleu,wieting-etal-2019-beyond,sellam2020bleurt,zhangbertscore,gehrmann2021gem}. 
Because these metrics require comparisons with reference utterances, which are not  available at generation time, we 
 define referenceless versions based on pseudo-references, S$_{pseudo}$, created from the input DAs \citet{juraska2022diversifying}.
For any MR, we create its S$_{pseudo}$ by omitting the slot names and the DA name and then concatenating the categorical attribute values with spaces between them, and converting  boolean attributes, such as {\sc has\_multiplayer} $=$ no, into phrases using  the attribute name,  with a negation when needed, e.g. ``no multiplayer''. For example,  S$_{pseudo}$ for the MR at the top of Table~\ref{tab:ex_video_game_dataset-3DAs-sameMR} would be ``Call of Duty: Advanced Warfare excellent Sledgehammer Games M for Mature".  Pseudo-references are available at generation time, so we use them to calculate
pseudo-metrics for semantic accuracy and use them in ranking.  
\citet{juraska2019viggo} shows that the {\it relative}  differences of these pseudo-metrics  distinguish errorful  NLG utterances from correct ones. 
%Our prompt styles defined below then add the DA to S$_{pseudo}$  in different ways. 

\vspace{0.1in}
\noindent{\bf Fluency.}  Recent work suggests that the probability P(S) of a generated output S  according to an LLM is a good automatic and referenceless  measure of fluency \cite{kann2018sentence,suzgunetal2022}. We thus adopt
 P(S) to measure fluency, and use GPT-2 to calculate P(S). 
 %We have observed that, in general,  NLG outputs from LLMs do not suffer from problems of fluency. However outputs from smaller LMs such as GPT-Neo 1.3B or GPT-J 6.0B are often less fluent.

\vspace{0.1in}
%\subsection{Ranking}
\label{ranking-funcs}
\noindent{\bf Ranking.}   The ranking functions in Table~\ref{ranking-functions-table} 
aim to select NLG outputs that  maximize DA accuracy, semantic accuracy, and fluency.   Ranking function RF1   scores each candidate according to  Equation~\ref{likelihood}.
%, however if the predicted DA does not match the  DA in the input MR, the probability is set to 0. 
\begin{table}[htb]
    \small
  	\centering
  	\def\arraystretch{1.5}
    \begin{tabular}{p{0.7\linewidth}}
    \rowcolor{light-gray}
        \hline \textbf{RF1: DAC * SACC * P(S)} 
     \\
     \hline
      %\rowcolor{light-gray}
    	\textbf{RF2: DAC * SACC * pBLEU * P(S)} 
     \\
     \hline
      \rowcolor{light-gray} 
     \textbf{RF2\textsubscript{DA}: DAC | SACC |  pBLEU | P(S)} \\
        
        \hline
    	%\rowcolor{light-gray}
    	\textbf{RF3:  DAC * pBBLEU * P(S)} \\
        \hline
        	%\rowcolor{light-gray}
    \rowcolor{light-gray}	\textbf{RF4: pBBLEU} \\
    \hline
        	%\rowcolor{light-gray}
    \textbf{RF5: pBLEU} \\  \hline
     \end{tabular}
    \caption{Ranking functions\label{ranking-functions-table}. DAC = probability of the correct DA using a classifier. SACC = semantic accuracy using domain-specific SACC scripts. P(S) = LM probability as a measure of fluency. pBBLEU = pseudo-Beyond-BLEU to measure  semantic accuracy. pBLEU = pseudo-BLEU as a baseline. }
    \vspace{-0.1in}
\end{table}

After a qualitative analysis of the ranking outputs from RF1 on pilot data, we developed ranker  RF2  and RF2$_{DA}$ in Table~\ref{ranking-functions-table}.  
Our analysis revealed that the SER scripts often do not  detect hallucinations, but  
 pBLEU appeared to detect some hallucinations, so we add pBLEU to RF2. Ranking function RF2$_{DA}$  prioritizes one metric at each step, as represented by | in RF2$_{DA}$, enforcing DA correctness as more important  for dialogue than perfect SACC. Matching DA candidates are preferred, but if no candidates match the required DA, the DA class {\it other} is preferred, or otherwise, all $k$ candidates are selected. The second step selects candidates with the highest SACC. The third step aims to remove candidates with hallucinations by choosing the highest pBLEU outputs. The final step selects outputs with the highest fluency (P(S)).
%Another advantage of ranking by   $RF2_{DA}$ is that it avoids computing all automatic metrics for each utterance, as the space of outputs is drastically reduced at each step. 

%\begin{table}
%\centering
%\small
%\vspace{-.2in}
%\begin{tabular}{|l|c|c|c|}
%\hline
%\textbf{Measure} & \textbf{ViGGO} & \textbf{Laptop}  & \textbf{TV}     \\ \hline
%pBLEU &0.08 & -0.12 &   0.05\\ \hline
%pBBLEU & 0.52 & 0.32  &  0.45   \\ \hline
%pBLEURT & 0.38 & 0.17 &  0.26  \\ \hline
%pBERT precision & 0.33 & 0.14 & 0.36  \\ \hline
%pBERT recall & 0.03 &  -0.06 &  0.14    \\ \hline
%pBERT F1 & 0.20 & 0.04 & 0.26  \\ \hline
%\end{tabular}
%\end{wrapfigure}
%\vspace{-.2in}
%\caption{\label{tab:correlations} Pearson correlation between SACC and common semantic preservation measures when applied to pseudo-references. All correlations
%are statistically significant at p $<$ 0.001 .}
%\vspace{-0.1in}
%\end{table}

So far RF1, RF2 and RF2$_{DA}$ all use the domain-specific SACC score 
for measuring semantic accuracy.  To define a domain-independent ranking function, we calculate the
correlation of SACC with  pBLEU, pBBLEU, pBERT, and pBLEURT, defined in Section~\ref{SACC-sec}, on sample model outputs. See Table~\ref{appendix-tab-correlations} in Appendix~\ref{appendix-semantic-accuracy-pseudo}. The results show that pBBLEU~\cite{wieting-etal-2019-beyond} has the highest correlation across all three domains with  0.52 for Viggo,
0.32 for Laptop and  0.45 for TV. We thus define  RF3
by replacing SACC in RF1 with pBBLEU. We then define RF4 as pBBLEU alone, so we can compare our novel ranking functions to pBBLEU. Finally, as a baseline reflecting the fact that previous work uses BLEU as a single measure of goodness for NLG, we define R5 as pBLEU.

\section{Experimental Overview}
\label{exp-meth-sec}

%\subsection{Prompt Formats}

%manual template engineering 
 %prompts is to manually create intuitive templates based on human introspection.
 % discrete prompts 
% There has also been criticism that models don't differentiate against the prompts given, so certain instructive prompts \cite{webson-pavlick-2022-prompt}.
% Prompt-based \cite{madotto_promptds}
%explanation as to why we created different prompt formats 

\begin{figure} [t!h!]
    \centering
    \small
    \includegraphics[width=\columnwidth]{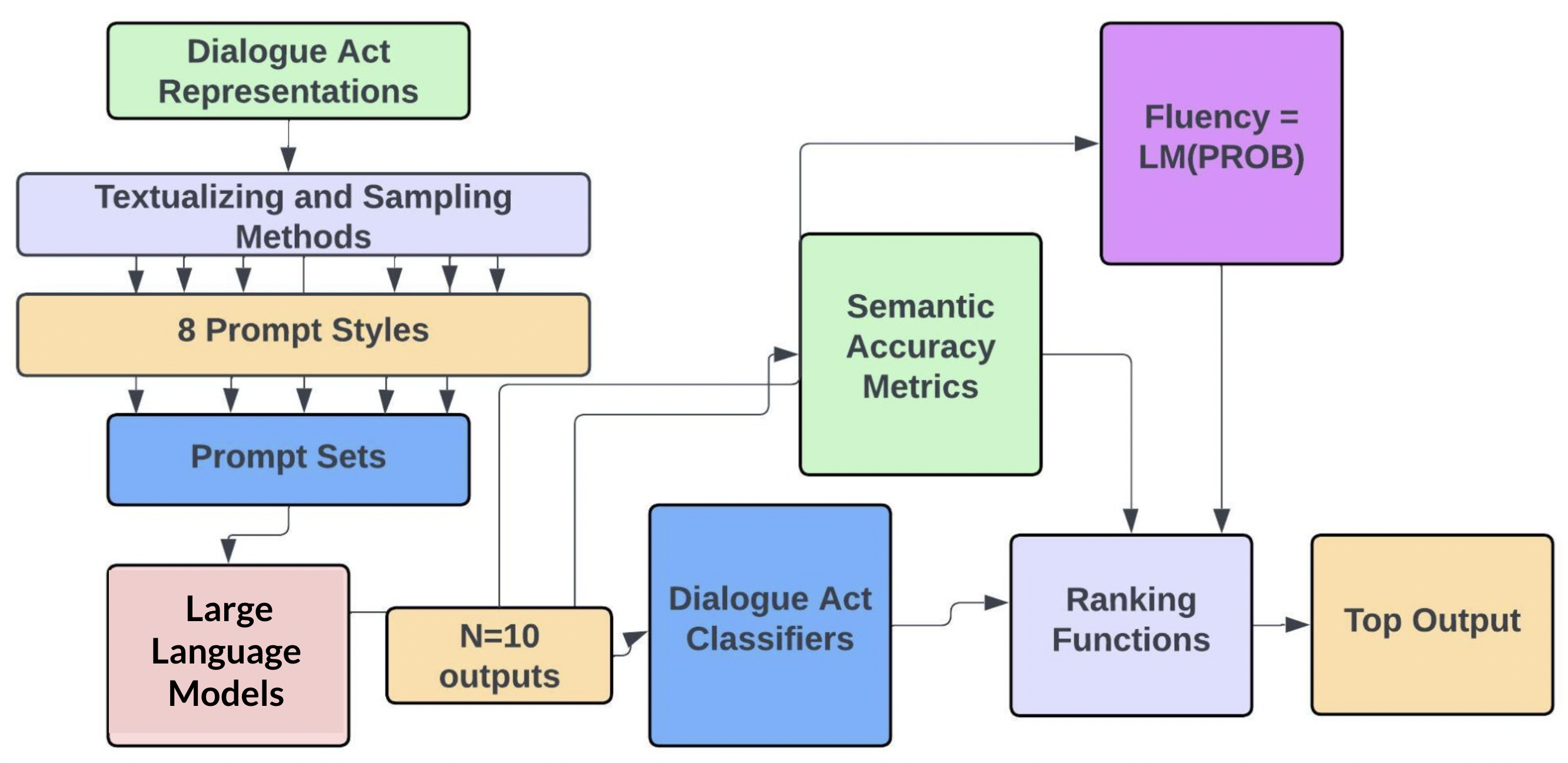}
    \caption{Experimental Architecture}
    \label{fig:exp-arch}
\end{figure}

Figure~\ref{fig:exp-arch} provides an overview of the experimental architecture.  Given a set of DA representations for a domain, we sample prompt examples from the original training sets while varying the number of samples. We then textualize the DA representations in the sample to look more similar to the LLMs free-text training data. The samples are then fed through the 8 prompt formats in Table~\ref{tab:prompt-formats}.
We apply this method to the ViGGO, Laptop and TV domains and
utilize the 6 ranking functions  in Table~\ref{ranking-functions-table}.
%We primarily experiment with the Jurassic Jumbo LLM but to test generalization, we test our best experimental parameters with GPT-3, GPT-Neo 1.3B and GPT-J 6.B.
%for a total of 31 experimental combinations.

\vspace{0.1in}
\noindent{\bf Prompt Formats.} LLMs %such as Jurassic and GPT-3 
are typically trained on  far more monologic data than dialogue, and will have rarely, if at all, seen examples of data-to-text NLG \cite{brown2020language, raffel2020exploring, devlin2018bert}.  While there are LLMs trained on dialogue such as DialoGPT \cite{zhang2020dialogpt}, and semantically-controlled dialogue data such as KGPT \cite{chen2020kgpt},  and SC-GPT \cite{peng-etal-2020-shot}, there are clear benefits to using a general LLM. %such as Jurassic. 
Previous work also shows that without specific  dialogic data, many LLMs do well on NLG for dialogue \cite{soltan2022alexatm}. Here, we test the hypothesis that performance can be improved by using prompt formats that make the data-to-text task look more like the LLM's textual training data. 

\begin{table}[ht!b]
%\begin{footnotesize}
%\begin{small}
\small
\begin{tabular}
%{@{} p{2.2cm}|p{4.7cm}}
{p{1.8cm}|p{5.1cm}} 
%\toprule
\hline
\textbf{Prompt ID} & \textbf{Prompt Template}  \\ \hline \hline 
\rowcolor{light-gray}{\sc tst vanilla}& Here is a text: ``$s_{pseudo}$". Here is a rewrite of the text which is a(n) $d$ dialogue act: ``$r_{text}$"  \\ \hline
{\sc tst} &  Here is a text: ``$s_{pseudo}$". Rewrite it \\
{\sc dialogue}&  to be a(n) $d$ dialogue act: ``$r_{text}$" \\ \hline
\rowcolor{light-gray}{\sc tst} & Here is a text: ``$d_r \ s_{pseudo}$". \\
\rowcolor{light-gray}{\sc paraphrase} & Here is a paraphrase of the text: ``$r_{text}$"  \\ \hline

{\sc definitional } & description of $<d>$: $D^{d}$.   \newline
Data:  $d = yes \ | \ sa^1 = v^1 .. sa^n = v^n$   \newline
Data to Text for $<d>$: $r_{text}$    \\ \hline
\rowcolor{light-gray} {\sc paraphrase } & $ d_r \ s_{pseudo}   \newline
     r_{text}$    \\ \hline
{\sc dialogic}  &  $ d_r \  s_{pseudo}$ \\
  &                $r_{text}$  \\   \hline
\rowcolor{light-gray} {\sc pseudo} & $ d \ s_{pseudo}   \newline
     r_{text}$     \\ \hline
 {\sc s2s} &  $d = yes \ | \ a^1 = v^1  .. a^n = v^n$ \newline
 $r_{text}$ \\ \hline

\end{tabular}
%\end{small}
\vspace{-0.1cm}
\caption{Prompt IDs and templates.  Instantiations of each  template  are given in Table~\ref{tab:full-prompt-formats} in the Appendix. \label{tab:prompt-formats}}
\end{table}

Table~\ref{tab:prompt-formats} shows the 8 prompt templates, with full instantiations in the Appendix in Table~\ref{tab:full-prompt-formats}.
The templates vary the representation of the DAs and their attributes.
We  represent the DA directly by its name~$d$, or convert the DA to a sentence starter d$_r$ such as ``I~recommend''.
The attributes of the DA constitute a set  $a = {a^1, a^2,...,a^n}$, each with a value in $v$ where $v= {v^1,v^2,...,v^n}$. The attributes can be  represented directly or using a textual pseudo-reference $s_{pseudo}$, as described in Section~\ref{SACC-sec}.
The reference text $r_{text}$ then varies the representation of the  DA and the attributes. 

Prompts  TST Vanilla,  TST Dialogue, and  TST Paraphrase of  Table~\ref{tab:prompt-formats}  treat data-to-text generation as a  textual style transfer (TST) task, 
where each DA is a style, and the prompt provides
instructions, e.g., ``Rewrite it to be a suggest dialogue act'' \cite{reif-etal-2022-recipe, suzgunetal2022}.  TST Vanilla and TST Dialogue represent the MR as its pseudo-reference $s_{pseudo}$, while TST Paraphrase prefixes the sentence starter d$_r$ for the DA to $s_{pseudo}$. 

We also define a Definitional prompt with definitions of the DAs, represented as $D^{d}$,   based on the instructions given to crowdworkers when ViGGO  was collected,  inspired by previous work providing
slot descriptions \cite{gupta-descriptions}.
 
The Paraphrase  prompt is based on the fact that producing paraphrases is a common task.
This prompt  rewrites the DA 
as a first-person sentence starter, e.g., ``I suggest'' for the {\it suggest} DA.
The  Dialogue Response prompt is similar, but  mimics  a request and its response, with sentence starters written as requests, e.g.,
%``can you suggest a game Worms:Reloaded Steam?" for the {\it suggest} DA and 
``can you recommend a game Worms:~Reloaded Steam?'' for the {\it recommend} DA.

To directly evaluate the benefit of instructions,   we also input the pseudo-reference  without instructions as a baseline (Pseudo), as well as input the commonly used S2S format which linearizes  the MR as a sequence of attributes and values \cite{soltan2022alexatm,wen2015semantically,harkous-etal-2020-text}.

\section{Results}
\label{results-sec}

\noindent{\bf Experimental Roadmap.}   We first experiment with  ViGGO  over all the experimental settings from Section~\ref{exp-meth-sec} using Jurassic-1 Jumbo, a 175B auto-regressive 
transformer-based LLM with a  different depth-width tradeoff  than GPT3 \cite{levine2020depth,lieberetal21}.
All experiments set top P $=$ 1, and T = 0.7 based on pilot experiments. %We also vary whether we provide multiple types of DAs in the
%prompts (Per-DA), or a single type of DA  (Specific-DA). 
We compare prompting  to few-shot fine-tuning 
 using 5, 25, 50 and 100 examples per DA sampled from the training data. We  
 test the 8 prompt formats in Table~\ref{tab:prompt-formats} with 1, 5 or 10 prompt examples. Our focus is DA control, so we create a ViGGO test set with 40 instances per DA (360 total). We look-ahead to see  which ranking function performs best for ViGGO and use that for the results  in Table~\ref{tab:performance-all}. 
  
We then test the best settings  from ViGGO on the Laptop and TV corpora \cite{wen2015semantically} with results in Table~\ref{tab:laptop-TV-performance-all}. We  compare ranking function performance across all domains in Table~\ref{tab:ranking-results}, and demonstrate the improved performance of our ranking functions compared to simply using BLEU. We then test for generalization with  additional LLMs: we select the top three prompt settings, and test of GPT-Neo as a smaller LLM, and GPT-3 and ChatGPT as instruction-tuned LLMs, and compare them to Jurassic-1, for all three domains. These results are shown in Table~\ref{tab:other-LLMs-results}. Table~\ref{table:sota-comparison} then compares our best performance  to recent SOTA  results for both fine-tuning and few-shot fine-tuning on ViGGO, Laptop and TV. Finally we report the results of our human evaluations. We make the DA classification models, the prompts and their instantiations, and the model outputs for all  experiments available.\footnote{ \scriptsize{\url{https://github.com/anon-nlp-1234/da-nlg}}}  %We also make available ViGGO model outputs that have been hand-labelled for DA that we use in our DA classifier.

\begin{table}[h!tb]
\centering

%\scriptsize
%\scalebox{0.4}{
%\resizebox{\textwidth}{!}{%
\begin{small}
\begin{tabular}{|p{.84in}|p{.15in}|p{.38in}|p{.38in}|p{.38in}|}
\hline
{\textbf{ID}} &
  {\textbf{N}} &
  {\textbf{PERF }} &
{\textbf{SACC}} &
 {\textbf{DAC}} 
 \\ 
   \hline
   \hline
\rowcolor{light-gray} \multicolumn{5}{|c|}{\textbf{ Few-Shot Fine-Tuning Experiments } }\\
FTune      5-per & 45 &38.88 & 85.71 & 54.44   \\
FTune      25-per & 225 &62.22 & 92.19   & 79.72  \\
FTune     50-per &450 &71.94 & 96.43   & 79.44  \\
FTune     100-per & 900 & \bf 78.61 & \bf 97.74  & \bf 80.56 \\

         \hline
         \hline
% \rowcolor{light-gray} \multicolumn{5}{|c|}{\textbf{ Prompt-Based In-Context Per-DA Experiments} }\\
        
% 5-per  DA     &45          & 60.0 & 83.6 & 92.0  \\ \hline % ROW 7 in sheet
% 2-per DA  &  18 & 61.0  & 83.2 & 91.4  \\ \hline  % ROW 19 in sheet
% 2-per DA + Def & 18 & 64.0  & 84.0 & 98.6  \\ \hline % ROW 18 in sheet
%    \hline
    	\rowcolor{light-gray} \multicolumn{5}{|c|}{\textbf{Prompt Styles and Samples Experiments} }\\
\rowcolor[HTML]{CCCCCC} 
TST Vanilla &
  10 &
  \textbf{85.56} &
  \textbf{94.73} &
  \textbf{100.00} 
  \\ \hline
\rowcolor[HTML]{CCCCCC} 
TST Dialogue &
  10 &
  83.89 &
  94.17 &
  100.00 
  \\ \hline
  \rowcolor[HTML]{CCCCCC} 
TST Paraphrase &
  10 & 83.90
   & 94.20
   & 100.00
  \\ \hline

\rowcolor[HTML]{CCCCCC} 
Definition (each) &
  10 &
  76.94 &
  91.16 &
  100.00
    \\ \hline
\rowcolor[HTML]{CCCCCC} 
Definition (top) &
  10 &
  82.22 &
  93.51 &
  100.00 
\\ \hline
\rowcolor[HTML]{CCCCCC} 
Paraphrase &
  10 &
  77.78 &
  92.10 &
  100.00 
   \\ \hline
\rowcolor[HTML]{CCCCCC} 
Dialogic &
  10 &
   77.22&
   91.53 &
  100.00 
   \\  \hline
 \rowcolor[HTML]{CCCCCC} 
Pseudo &
  10 &
  75.83 &
  94.17 &
  100.00 
  \\ \hline
  \rowcolor[HTML]{CCCCCC} 
S2S &
  10 &
  70.56 &
  86.45 &
  100.00 
  \\   
   \hline \hline
\rowcolor[HTML]{EFEFEF} 
TST Vanilla &
  5 &
  80.56 &
  92.57 &
  99.72        
  \\ \hline
\rowcolor[HTML]{EFEFEF} 
TST Dialogue &
  5 &
  \textbf{83.61} &
  \textbf{93.88} &
  \textbf{100.00} 
           \\ \hline
\rowcolor[HTML]{EFEFEF} 
TST Paraphrase &
  5 & 80.20
  & 92.60
  & 99.70
          \\ \hline
\rowcolor[HTML]{EFEFEF} 
Definition (each) &
  5 &
  80.00 &
  92.66 &
  99.40         
 \\ \hline
\rowcolor[HTML]{EFEFEF} 
Definition (top) &
  5 &
  77.22 &
  91.25 &
  100.00       
 \\ \hline
\rowcolor[HTML]{EFEFEF} 
 Paraphrase &
  5 &
  70.83 & 
   89.71&
  100.00      
  \\ \hline
\rowcolor[HTML]{EFEFEF} 
Dialogic &
  5 &
  66.94 &
  88.34&
  99.10        \\  \hline
 \rowcolor[HTML]{EFEFEF} 
Pseudo &
  5 &
  52.22 &
  82.60 &
  85.56         
 \\ \hline
 \rowcolor[HTML]{EFEFEF} 
S2S &
  5 &
  66.67 &
  83.54 &
  99.72        
  \\  
  \hline \hline
TST Vanilla &
  1 &
  68.06 &
  86.64 &
  91.94      
  \\ \hline
TST Dialogue &
  1 &
69.17 &
  88.15 &
  93.30          
   \\ \hline
TST Paraphrase &
  1 & \textbf{72.20}
 &\textbf{89.80}  & 93.60
   \\ \hline
Definition &
  1 &
  63.89 &
  85.32 &
  \textbf{98.30}      
  \\ \hline
Paraphrase &
  1 &
  41.94 &
  75.14 &
  83.88 
    \\ \hline
Dialogic &
  1 &
   38.89  & 
  71.83 &
  82.30 
  \\ \hline
\end{tabular}
\end{small}
%}

%}
\caption{Results after ranking via RF2\textsubscript{DA} for ViGGO. N $=$ number of prompt examples. PERF = \% outputs that are perfect. SACC = semantic accuracy using SACC scripts. DAC = DA accuracy using a classifier. 
% See Table~\ref{tab:other-metrics-performance-all} in the Appendix for before/after results and stats on how ranking  significantly improves all metrics.  
}
\vspace{-0.1in}
\label{tab:performance-all}
\end{table}

%\subsection{Quantitative Results}
%

% One liner: DISCUSS BOTH WHOLE POOL/FIRST TEXT and after ranking 

% One liner: WHAT WE'RE MEASURING mention diversity metrics, performance on SACC and DA. corpus bleu/pseudo bleu

% One liner: CALCULATE CORRELATION COEFFICIENTS TO OTHER DIVERISTY METRICS AND HAVE ONE LINER , IF THIS IS CORRELATED WE ONLY MENTION VOCABULARY AND NO OTHER METRICS BECAUSE OF THAT 

% TABLE FOR AFTER RANKING + Results on First SEt

% TABLE FOR ALL RANKING FUNCTIONS ON THE BEST SETS

%\subsection{Few-Shot Fine-tuning}
\vspace{0.1in}
\label{few-shot-ft}
\noindent{\bf Few-Shot Fine-Tuning.} To compare prompting to 
 fine-tuning, we use the traditional linearized MR in the S2S format  and vary the number of training examples per DA in few-shot fine-tuning  from 5, to 25, to 50, to 100. 
The results in Rows 1-4 of  Table~\ref{tab:performance-all} show that, as expected, increasing the number of training examples  improves performance, with 100 examples per DA (900 overall) achieving a SACC of 97.74 after ranking. However, interestingly, the highest DAC performance is only 80.56, and the PERF score   (both perfect DA and perfect SACC) is only 78.61. 
Table \ref{tab:ftvsnumprompt} in the Appendix shows more detail, providing before and after ranking
performance for fine-tuning. 
%With just 5 prompts, ranking a gives similar performance as that of 100 prompts without ranking.
%The second section of Table \ref{tab:performance-all} also shows for the same number of samples, prompting in Per-DA format ($n=5$ prompts per DA) achieves higher DA accuracy by a large margin (92.00 vs 54.44) over fine-tuning. 
Overall, the results affirm previous findings that few-shot prompting beats few-shot fine-tuning \cite{le-scao-rush-2021-many}. 
%Row XXX  of Table~\ref{tab:performance-all} shows that prompting on just a single definitional prompt beats the fine-tuned model of 225 data points ($n=25$) 

% %\subsection{Multiple Prompts Per DA}
% \vspace{0.1in}
% \noindent{\bf Multiple Prompts Per DA.}
% The second section of Table~\ref{tab:performance-all} shows results for
% the Per-DA sampling experiments. 
% Per-DA sampling uses prompts with examples from all DAs in order to test whether the LLM  benefits from learning generalizations over the 9 ViGGO DAs. So a 5-prompt Per-DA experiment has a total of $5 \times 9 = 45$ examples.  

% Overall Table \ref{tab:performance-all} shows that performance is worse when multiple types of DAs are provided in the examples, rather than benefiting from different types of examples. Table \ref{tab:performance-all} also shows  that 2 examples per DA performs on par with 5 examples per DA. 

% The Multi-DA  experiments also test the Definitional-Prompt style shown in Table~\ref{tab:prompt-formats}. The token limit of Jurassic (2048 tokens) supports 2 prompt examples 
% with definitions Per-DA. Interestingly, this performs best of the Per-DA experiments on all metrics, with DA accuracy improving from 91.4\% to 98.6\%.

%\subsection{Specific-DA Prompt Styles.} 
\vspace{0.1in}
\noindent{\bf Prompt Styles.} All experiments  provide  examples for a single DA and then generate  that DA, while varying the prompt style and the number of examples.
%{\color{red} since multi DA removed, this doesn't need to keep talking about specific}
%The second section of Table~\ref{tab:performance-all} shows results of experiments using 1, 5 and 10 Specific-DA prompts. Per-DA experiments could be run with at most 5 examples per DA, so  we  compare Specific-DA sampling to  Per-DA sampling for 1 and 5 samples. As shown in Table~\ref{tab:performance-all}, the best 5 Per-DA performance for PERFECT is 60\%, whereas the best 5 Specific-DA performance is 83.6\%. 
%Consider the performance of different prompts in the second section of Table~\ref{tab:performance-all}. 
%Ranking always significantly improves all metrics. We calculate \textit{Before Ranking} by averaging all metrics over the entire 3600 set to take into account randomness. \textit{Before Ranking} produces lower averages than \textit{After Ranking} where for Corpus SACC Before averaged 60\% and After averaged 83\%, DAC Before averaged 68\% and After averaged 96\%, and DA+SACC Before averaged 32\% and After averaged 62\%.  
% When evaluating results for \textit{After Ranking} for all prompt styles in table \ref{tab:specific_performance}, we find that the more number of prompts the higher the DA accuracy and SACC. 
The TST format provides N examples using one of the TST prompts in Table~\ref{tab:prompt-formats}. The  Definitional (each) format,  for 10 prompts, provides  10 triplets of (definition, MR, text). For Definitional (top), the definition is mentioned once before all the MRs and examples, so for 1 prompt, there is no difference between \textit{top} and \textit{each}.

We first notice in Table~\ref{tab:performance-all} that the PERF score  improves with the number of prompt examples, from 1 to 5 to 10 for all the prompt styles, with
TST Vanilla, TST Dialogue, and TST Paraphrase, which provide the MR as text and include instructions  (see Table \ref{tab:prompt-formats}) consistently performing the best overall. 
TST Vanilla-10 performs signicantly better than the other TST styles with 10 examples (p $<$ .01),  but TST Dialogue is the best for 5 examples and TST Paraphrase is the best for 1 example. 
The Definitional, Paraphrase and Dialogic formats all perform significantly worse than the TST formats, but interestingly the Definitional format gets the highest DAC with only 1 example perhaps showing the advantage of explicit definitions in PBL.

The Pseudo and S2S  prompt styles  are  baselines, and only reported for the 5 and 10 example settings. Both baselines indicate the benefits of instructions.
The S2S 10 performance is the worst for 10 examples, and the Pseudo  performance is the worst for 5 examples. It is worth noting that the poorly performing S2S representation is
commonly used in both fine-tuning and PBL~\cite{soltan2022alexatm,wen2015semantically,harkous-etal-2020-text}. 

\begin{table}[h!tb]
\centering
%\scriptsize
%\scalebox{0.4}{
\resizebox{.48\textwidth}{!}{%
\begin{small}
\begin{tabular}{|p{.35in}|p{.55in}|p{.15in}|p{.38in}|p{.38in}|p{.38in}|}
\hline
{\textbf{Domain}} &
{\textbf{ID}} &
  {\textbf{N}} &
  {\textbf{PERF }} &
{\textbf{SACC}} &
 {\textbf{DAC}} 
 \\ 
   \hline
   \hline
%\rowcolor{light-gray} \multicolumn{5}{|c|}{\textbf{Laptop } }\\
Laptop & TST Van.    & 10 & 80.95 & 95.90 & 100.00 \\ \hline
TV & TST  Van.   & 10 & 98.85 & 99.76  & 100.00 \\ 
\hline
\end{tabular}
\end{small}
}
\caption{Results for Laptop and TV for TST 10 using RF2\textsubscript{DA}. N $=$ number of examples. PERF = \% outputs that are perfect. SACC = semantic accuracy using SACC scripts. DAC = DA accuracy using a classifier. 
%See Table~\ref{tab:other-metrics-performance-all} in the Appendix for before/after results and stats on how ranking  significantly improves all metrics.  
}
\vspace{-0.1in}
\label{tab:laptop-TV-performance-all}
\end{table}

We then take the best performing prompt (TST Vanilla) and experiment with TV and Laptop. The results
are shown in Table \ref{tab:laptop-TV-performance-all}.  RF2$_{DA}$ performs the best for both Laptop and TV so these results are ranked 
with RF2$_{DA}$. Interestingly, TV has the highest PERF and SACC seen so far, while Laptop also has a higher SACC than any ViGGO setting, suggesting that it is easier to achieve high performance with Laptop and TV than ViGGO.

\begin{table}[h!tb]
\centering
\resizebox{.48\textwidth}{!}{%
\begin{small}
\begin{tabular}{|c|c|c|c|c|c|}
\hline
\textbf{RF} & \textbf{Terms} &  PERF  &  SACC   &   DAC & BLEU\\
\hline
\rowcolor{light-gray} \multicolumn{6}{|c|}{\textbf{ ViGGO} }\\
RF1 & {\scriptsize \bf DAC, SACC, P(S)} &  79.17 &91.82&   99.72 & 38.41\\
RF2 & {\scriptsize \bf DAC, SACC,  pBLEU, P(S)}  &78.33& 91.72&   99.00 & 38.67 \\
RF2\textsubscript{DA}& {\scriptsize \bf DAC, SACC, pBLEU, P(S)}  & \textbf{85.56}& \textbf{94.73}&   \bf 100.00 & 40.08\\
RF3 & {\scriptsize  \bf DAC,  pBBLEU, P(S)} &   62.78 & 84.38 & 100.00 &  \bf 49.87 \\
RF4 & {\scriptsize{ \bf pBBLEU}} &   60.55 & 91.63 & 77.78 & 42.82\\  
RF5 & {\scriptsize{ \bf pBLEU}} &   44.22  & 81.66 & 75.28 & 40.08\\ 
\rowcolor{light-gray} \multicolumn{6}{|c|}{\textbf{ TV }} \\
RF1 & {\scriptsize \bf DAC, SACC, P(S)} &    85.40 &96.86 &   100.00 & 72.55\\
RF2 & {\scriptsize \bf DAC, SACC,  pBLEU, P(S)} &  88.19 & 97.43&  100.00  & 72.55 \\
RF2\textsubscript{DA}& {\scriptsize \bf DAC, SACC, pBLEU, P(S)} &  \textbf{98.85}& \textbf{99.76}&   \bf 100.00 & 60.51\\
RF3 & {\scriptsize  \bf DAC,  pBBLEU, P(S)} &   73.96& 93.87 &  100.00 & \bf 72.89\\
RF4 & {\scriptsize{ \bf pBBLEU}} & 90.14  & 97.88  & 99.71 & 60.51\\    
RF5 & {\scriptsize{ \bf pBLEU}} &  63.45  & 91.50 & 99.57 & 66.71\\ 
\hline
%\textbf{RF} & \textbf{Terms} &  PERF  &  SACC   &   DAC & BLEU\\
\rowcolor{light-gray} \multicolumn{6}{|c|}{\textbf{ Laptop} }\\
RF1 & {\scriptsize \bf DAC, SACC, P(S)} &   49.25 & 86.70&  100.00 &  61.24\\
RF2 & {\scriptsize \bf DAC, SACC,  pBLEU, P(S)} &  57.29 & 89.47 &  100.00 & 59.39\\
RF2\textsubscript{DA}& {\scriptsize \bf DAC, SACC, pBLEU, P(S)} &  \textbf{80.95}& \textbf{95.90}&   \bf 100.00 & \bf 61.36\\
RF3 & {\scriptsize  \bf DAC,  pBBLEU, P(S)} &  35.55 & 80.41 & 100.00 &  45.03\\
RF4 & {\scriptsize{ \bf pBBLEU}} &   61.79 & 90.97  & 98.88 & 36.32\\    
RF5 & {\scriptsize{ \bf pBLEU}} & 42.38   &  84.25 & 97.77 &  \bf 61.36 \\ \hline                                                          
\end{tabular}
\end{small}
}
\caption{Ranking functions performance. }
\label{tab:ranking-results} 
\end{table} 

%\subsection{Ranking Functions}
\vspace{0.1in}
\noindent{\bf Ranking Functions.} Our results show that our overgenerate-and-rank method has a huge effect on performance as compared to taking the first output from the model. Section~\ref{appendix-before-after} in the Appendix provides more detail, e.g. showing for Viggo, across all the experiments, \textit{Before Ranking} has an average SACC of 65.29\% versus an \textit{After Ranking} average of 86.82\%, while DAC has an almost a 30\% increase with a \textit{Before Ranking} average of 62.11\%, and an \textit{After Ranking} average of 91.04\%. 

Table~\ref{tab:ranking-results} compares the 5 ranking functions from Section~\ref{ranking-funcs} on all three domains for the best prompt so far: TST Vanilla 10.  The differences between RF1 and RF2 (addition of pBLEU)  are not significant for ViGGO, but are significant for TV (t-test, p $<$ 0.001) and Laptop (t-test, p $<$ 0.001), with Laptop  improving from 49.24 PERF to 57.29 PERF.  Note that in all domains ranking by RF2\textsubscript{DA} results in significantly higher performance across all metrics  (t-test, p $<$ 0.001): {\bf prioritizing  DA  correctness results in higher SACC and  higher PERF}.

Table~\ref{tab:ranking-results} also shows that replacing SACC with pBBLEU in RF3 results in a clear drop in performance. As shown in Appendix Section~\ref{appendix-semantic-accuracy-pseudo} pBBLEU is the best performing pseudo-metric overall, but there are clear advantages to the domain-specific SACC. Recent work explores automatic methods for training domain-specific  semantic fidelity classifiers, but these methods rely on large training corpora making them  difficult to apply in few-shot settings \cite{harkous-etal-2020-text,batra-etal-2021-building}.

The baseline RF4 with only the pBBLEU term  performs surprisingly well in SACC across all three domains, suggesting that it might be worth examining further combinations of BBLEU with DAC.

\begin{table}[ht!b]
\centering
\resizebox{.48\textwidth}{!}{%
\begin{small}
\begin{tabular}{|c|c|c|c|c|c|}
\hline
MODEL & PROMPT & PERF  &  SACC   &   DAC & BLEU\\
\hline
\rowcolor{light-gray} \multicolumn{6}{|c|}{\textbf{ ViGGO} }\\
ChatGPT & TST 10  & 98.89 &	95.58	 &99.44	 &45.05 \\
ChatGPT & TST 5	& 94.72 &99.34	 &96.67 &	40.88\\
ChatGPT  & Def 10	& 98.89 & \bf 100.00	 &100.00 &	42.40	\\
~\colorbox{cyan!30}{ChatGPT VO} & Def 10	& 95.28 & 99.85	 & 95.83 &  14.79\\	\hline											
GPT 3  &TST 10	& 95.00 &98.49	 &98.33 &	40.26	\\																
GPT3  &TST 5	&  95.28 &98.31 &	98.89	 & \bf 54.11		\\														
GPT3  &Def 10 &  \bf 99.44 &	 99.81 &	\bf 100.00	 &42.75		\\	
~\colorbox{cyan!30}{GPT3 VO} &Def 10 &  95.28  &	99.83  &	95.55	 &  9.55\\ \hline								

Jurassic  &TST 10	& 85.56 & 94.70	 &  100.00 & 40.08 \\ 												
Jurassic  &TST 5	&  83.61 &	93.88  & 100.00 & 32.54	\\													
Jurassic &Def 10 & 82.22 &	93.51 & 100.00 & 15.77 \\							\hline	

%GPT 6B  & TST 10 &&	92.57 &	38.89	 &23.86\\
%GPT 6B  & TST 5 dial &&	90.03 &	96.11 &	39.8\\
%GPT 6B  & Def 10 &&	75.7	 &93.06	 &23.36\\
GPT NEO 1.3B  & TST 10	 & 17.78 &85.32	&35.56 &	25.25\\
GPT NEO 1.3B  & TST 5 dial  & 64.17 &	86.74	 &94.72  &	43.47\\
GPT NEO 1.3B  & Def 10  &35.56 &	78.27 &	81.94  &	15.44	\\ \hline
\rowcolor{light-gray} \multicolumn{6}{|c|}{\textbf{TV} }\\ 
%  MODEL & PROMPT & PERF  &  SACC   &   DAC & BLEU\\
ChatGPT & TST 10  &  98.00 &	99.57	 &99.93	 &45.98 \\
ChatGPT & TST 5	& 91.23 &98.14	 &100.00 &	38.22\\
ChatGPT  & Def 10	&  98.00 &99.30 &99.64 &	50.97	\\	
 \hline									

GPT 3  &TST 10	& \bf 99.57 &  99.91	 & \bf 100.00 &	57.92	\\												
GPT3  &TST 5	& 99.07 &99.81 &	100.00	 &71.80		\\														
GPT3  &Def 10 & 99.22 &	 \bf 99.94 & 100.00	 &  73.81	\\
 \hline											
Jurassic  &TST 10	&  98.85 &   99.76&  100.00 & 60.51	\\ 												
Jurassic  &TST 5	&  91.80 &	98.26  & 100.00 & \bf 74.73	\\													
Jurassic &Def 10 & 95.01 &	98.94 & 100.00 &	73.66\\							\hline	

GPT NEO 1.3B  & TST 10	 & 83.15 & 96.37	& 100.00 &	66.28\\
GPT NEO 1.3B  & TST 5 dial  & 50.78 &	93.15	 & 73.93 &	31.95\\
GPT NEO 1.3B  & Def 10  & 15.74 &	78.61 &  65.88 &	19.29	\\ \hline											
  \rowcolor{light-gray} \multicolumn{6}{|c|}{\textbf{ Laptop} }\\   
%  MODEL & PROMPT & PERF  &  SACC   &   DAC & BLEU\\
  ChatGPT & TST 10  &  \bf 97.08 &  99.47 	&  99.58	 	 &  41.45\\
ChatGPT & TST 5	& 85.95  &	97.19 & 99.43 &	23.36\\
ChatGPT  & Def 10	& 67.54  &	90.37  &  99.92 &	36.00\\ 	\hline									
GPT 3  &TST 10	& 84.79 & \bf 99.91	 & \bf 100.00 & 33.20	\\ 														
GPT3  &TST 5	&  94.79 &	97.14  & 100.00 & 32.41	\\													
GPT3  &Def 10 & 81.45 &	92.54 & 100.00 &	\bf 85.40\\							\hline		

Jurassic  &TST 10	& 80.95 & 95.90	 & 100.00 & 61.36	\\ 												
Jurassic  &TST 5	&  81.55 &	96.10  & 99.81 & 12.94	\\													
Jurassic &Def 10 & 55.98 &	45.60 & 100.00 &	 29.12\\							\hline

GPT NEO 1.3B  & TST 10	 & 68.89 & 92.66	 & 100.00  & 46.21	\\	
GPT NEO 1.3B  & TST 5 dial  & 71.89 & 93.55	 & 100.00  & 19.49	\\	 
GPT NEO 1.3B  & Def 10   & 1.33 & 43.73  	 & 99.96 & 14.59	\\ \hline	
	
\end{tabular}
\end{small}
}
\caption{Experiments with additional LLMs, with the top three prompt settings, for ViGGO, Laptop and TV, using the RF2\textsubscript{DA}  ranking function.  We also tested here with the original ViGGO test set, with ChatGPT Def 10 and GPT-3 Def 10, with results shown in cyan, to facilitate comparison with previous work.
\label{tab:other-LLMs-results}} 
\end{table} 

Finally, the pBLEU baseline of RF5 reinforces work emphasizing the inadequacies of BLEU as a metric for NLG \cite{belz2008automatic,liu2016not,novikova2017we}. We report BLEU for comparison with related work, but Table~\ref{tab:ranking-results} clearly shows that the highest BLEU score doesn't correspond to the best PERF or SACC, and that even ranking with pBLEU (RF5) doesn't maximize BLEU.  RF5 gets the lowest PERF, SACC and DAC scores for ViGGO and TV, and  RF2${_DA}$ achieves the same  BLEU score, with much higher PERF, SACC and DAC for both ViGGO and Laptop. 

\vspace{0.1in}
\noindent{\bf Experiments with other LLMs}.
We also compare our results with Jurassic to other  LLMs.
We select the three best prompt settings, namely TST 10, TST 5, and Definitional Top 10, and experiment with ChatGPT and GPT-3 as large instruction-based models and GPT-Neo 1.3 as a small model. 

Table~\ref{tab:other-LLMs-results} presents the results.   Our primary metric is PERF with best PERF shown in bold. Note in the table that the highest PERF score does not necessarily correspond with the highest SACC or highest BLEU. Interestingly, GPT-3 performs slightly better than ChatGPT for both ViGGO and TV while ChatGPT performs best for Laptop. Both ChatGPT and GPT-3 perform significantly better than Jurassic across all three domains.    Table~\ref{tab:other-LLMs-results} shows that the Definitional prompt performs better than TST 10 with both ChatGPT and GPT-3 for Viggo, while TST 10 for TV was comparable to Definitional and performs the best for Laptop in terms of PERF. We add results here for the original ViGGO test set shown in cyan, which has a skewed distribution of DAs with  more long {\textsc inform} DAs, and which appears to be more challenging for DAC but not SACC. Finally, we see much worse performance  with GPT Neo, reinforcing results suggesting a model size threshhold for  PBL \cite{weiemergent}.

\noindent{\bf Comparison with SOTA}. Table~\ref{table:sota-comparison} compares our best results with recent work on the VIGGO, Laptop and TV corpora \cite{xu2021augnlg,du2022self,juraska-walker-2021-attention,kedzie2020controllable,harkous-etal-2020-text,peng-etal-2020-shot}. The related work either used fine-tuning or few-shot fine-tuning, rather than PBL. JW21, DT and K-McK are based on fine-tuning. 
SC-GPT, AUGNLG and ST-SA are all based on {\sc FewShotWoz}. In each case, we take the results exactly as reported in the  related work. These results are indicative only as  e.g. {\sc FewShotWoz} does not use the original RNN-NLG test set for Laptop and TV, which we use here. We created our own ViGGO test set to have equal numbers of each DA, but the original test set has many more long {\it inform} DAs. 

\begin{table}[t]
\centering
\resizebox{.48\textwidth}{!}{%
\begin{small}
	\begin{tabular}{|p{2.1cm}|p{0.7cm}p{0.7cm}|p{0.7cm}p{0.7cm}|p{0.7cm}p{0.7cm}|}
	\hline
	{\bf Model}    & \multicolumn{2}{c|}{\bf Laptop}   & \multicolumn{2}{c|}{\bf TV} & \multicolumn{2}{c|}{\bf ViGGO}  \\ 
				    & BLEU$\uparrow$ & ERR$\downarrow$    & BLEU$\uparrow$ & ERR$\downarrow$  & BLEU$\uparrow$ & ERR$\downarrow$  \\ \hline\hline
	\rowcolor{light-gray}     Ours    & 33.20 & \bf 0.08 &   \bf  73.81    &  \bf 0.06   &  14.79 &  \bf 0.15    \\   
             JW21    &  -- & -- & --& -- & \bf 53.60  &   0.46\\   
      \rowcolor{light-gray}         DT    &   &  & &  & \bf 53.60 & 1.68 \\ 
	     K-McK & --& -- &  -- & -- & 48.50 & 0.46    \\   
	\rowcolor{light-gray} SC-GPT      &32.73 & 3.39  & 32.95 & 3.38 & -- & --  \\
		
    AUGNLG-SC   & 34.32 &  2.83  &   34.99 &  5.53 & -- &  -- \\
    \rowcolor{light-gray} ST-SA & \bf 35.42 & 2.04 & 36.39 & 1.63 & --& --\\
	\hline
	\end{tabular}
 \end{small}
 }
\caption{
 Ours $=$ Our best model for each domain from Table~\ref{tab:other-LLMs-results} compared to recent SOTA results. Our VIGGO result is for  the ViGGO ORIGINAL test set.  JW21 $=$ SeaGuide \cite{juraska-walker-2021-attention}. DT = Data Tuner \cite{harkous-etal-2020-text}. K-McK $=$ \cite{kedzie2020controllable}. SC-GPT $=$ \cite{peng-etal-2020-shot}. AugNLG $=$ \cite{xu2021augnlg}. ST-SA $=$ \cite{du2022self}. We convert  SACC to SER, which other work calls ERR, and report BLEU, and ERR as in that other work.  Note that we use our best SACC score from Table~\ref{tab:other-LLMs-results} to select the row to include here, but this doesn't necessarily correspond to the best BLEU score or the best PERF score. 
\label{table:sota-comparison}}
\end{table}

%FT-GPT         && &28.83 & 11.82  & 33.73 & 9.28 &  &   \\ 
	%\rowcolor{light-gray}	

%\subsection{Human Ann}
\vspace{0.1in}
\noindent{\bf Human Evaluation.} Given the almost perfect performance reported in Table~\ref{tab:other-LLMs-results}, we conducted
a human evaluation to  check whether the outputs were indeed perfect (the right DA and the
correct semantics), and whether
there were any hallucinations. Two expert annotators hand-labelled 100 outputs from ChatGPT with
TST-10 Vanilla prompts. Amazingly, neither annotator found any outputs that weren't perfect  and neither did they find any hallucinations. They agreed 100\% on the results,
resulting in a Cohen's Kappa of 1.0.

We also  test whether our addition of pBLEU to RF2 has an effect on hallucinations, by testing in general whether pBLEU helps identify hallucinations. We annotate  
 hallucinations for ViGGO, by having 3 annotators label all 360 outputs for each ranking function (6*360) shown in Table~\ref{tab:ranking-results}. The number of hallucinations for RF1 was 34, RF2 was 19, RF3 was 26, RF4 was 40 
and RF5 was 14. We compared the mean number of hallucinations of ranking functions with pBLEU, namely RF2, RF2\textsubscript{DA}, and RF5  to those without, namely RF1, RF3 and RF4. We find that
the mean number of hallucinations of those with pBLEU is  31.67, while the mean number of those without is 19.67. 
This difference seems large, but the sample size is small and therefore it's not significant (t = 1.82, p = .14)

\section{Conclusion and Future Work}
\label{conc-sec}

Here we apply an overgenerate-and-rank NLG approach and and provide the first experiments using automatic ranking functions that optimize both DA and semantic accuracy in few-shot prompt-based NLG. 
We test and compare a combination of prompt formats,  sampling methods, and DA representations. We test prompts used  for textual style transfer (TST) by treating DAs  as styles to be controlled. We also create  novel prompts that provide definitions of DAs,  For completeness,  we fine-tune few-shot models  and compare them with  the  few-shot results.  
The results show that several prompting styles achieve perfect DA accuracy, and that few-shot methods can achieve semantic accuracy as high as 99.81\% with the right ranking function, while 100-shot fine-tuning  achieves 97.7\%, and performs much worse on DA accuracy (80.6\%). 

Our contributions include systematic experimentation with different ways of textualizing MRs, providing instructions to the LLM, and ranking outputs.   Our  results also show that formulating the data-to-text task as textual style transfer using pseudo-references yields the highest performance. We achieve SOTA semantic accuracy with only 10 prompt examples with our best prompt styles, and achieve the surprising results that a ranking function that prioritizes  DA  correctness results in higher semantic accuracy.

\noindent{\bf Limitations and Risks} One limitation arises from the challenges of prompt-engineering: it is impossible to tell whether another prompt format  could perform better, e.g. with smaller LLMs like GPT-Neo, where we get poor comparative results. Another limitation is the need for a high-accuracy DA classifier that works well on out-of-domain model outputs. We  address this limitation by releasing our classifiers. Another possible limitation is the use of the overgenerate and rank approach  in  real-time. In future work we plan to use the high quality (ranked) generated data, to fine-tune a smaller  real-time language model, without the need for overgeneration.
Another limitation arises from the comparison to few-shot fine-tuning -- there are  many ways to fine tune and many representations of the MRs, so it is possible that some other method of fine-tuning would lead to better fine-tuning results  \cite{liu2022few}. Our main goal here was to show that with a small-number of examples, using reasonable assumptions, few-shot fine-tuning performs worse than PBL.

A potential risk of  using LLMs is the possibility of disinformation, often called hallucinations. Control of  hallucinations is an active area of research. One of the challenges is that it is very difficult to automatically identify them. Here we experiment with ranking functions for better control of hallucinations, hand-label hallucinations and characterize them. Another potential risk of our work is that some of our dialogue acts like recommend and suggest could be used, in an application context, to persuade a user to buy something. In this context, it is even more important to ensure that the system is not providing false information to users.

\bibliographystyle{acl_natbib}

\clearpage
\newpage

\appendix

\section{Appendix}
\label{sec:appendix}

\subsection{Full Prompt Descriptions and Examples}
 Table~\ref{tab:full-prompt-formats} shows a sample instantiation for each prompt type and template. When this paper is accepted, we will provide
 all the prompt files and instantiated prompts for all experiments in our github: \url{https://github.com/aramir62/da-nlg}.

\begin{table}[b]
\centering
\footnotesize
\setlength{\tabcolsep}{4pt} % Adjust the column separation
\begin{tabular}{@{} p{2.2cm}|p{5.8cm} @{}}
\hline
\textbf{Prompt ID}  & \textbf{Example}  \\
\hline  
{\sc tst vanilla} & Here is a text: "Worms: Reloaded Steam". Rewrite of the text, which is a suggest dialogue act: "I bet you like it when you can play games on Steam, like Worms: Reloaded, right?" \\
\hline
{\sc tst dialogue} & Here is a text: "Worms: Reloaded Steam". Rewrite it to be a suggest dialogue act: "I bet you like it when you can play games on Steam, like Worms: Reloaded, right?" \\
\hline
{\sc tst paraphrase} & Here is a text: "I suggest Worms: Reloaded Steam". Paraphrase of the text: "I bet you like it when you can play games on Steam, like Worms: Reloaded, right?" \\
\hline
{\sc definitional } & Description of $<suggest>$: A question asking if your friend has any experience with a certain type (based on data) of video games. Use the name of the game in data with 'such as', 'like', etc. The response should consist of a single yes/no question. 
Generate diverse responses. \newline\newline\newline Data: suggest = yes $|$ name = Worms: Reloaded $|$ available\_on\_steam = yes. \newline Data to Text for $<suggest>$: 
 I bet you like it when you can play games on Steam, like Worms: Reloaded, right? \\
\hline 
{\sc paraphrase } & I suggest a game Worms: Reloaded Steam. \newline
I bet you like it when you can play games on Steam, like Worms: Reloaded, right? \\
\hline
{\sc dialogic}  & Can you suggest a game Worms: Reloaded Steam? \newline I bet you like it when you can play games on Steam, like Worms: Reloaded, right? \\
\hline
{\sc pseudo}& Suggest Worms: Reloaded Steam. \newline I bet you like it when you can play games on Steam, like Worms: Reloaded, right? \\
\hline
{\sc s2s} & suggest = yes $|$ name = Worms: Reloaded $|$ available\_on\_steam = yes. \newline I bet you like it when you can play games on Steam, like Worms: Reloaded, right? \\
\hline
\end{tabular}
\caption{Prompt IDs and Instantiation of each Prompt Template Type}
\label{tab:full-prompt-formats}
\end{table}

\subsection{Semantic Accuracy Pseudo Metrics}
\label{appendix-semantic-accuracy-pseudo}

\begin{table}[h]
\centering
\small
%\vspace{-.2in}
\begin{tabular}{|l|c|c|c|}
\hline
\textbf{Measure} & \textbf{ViGGO} & \textbf{Laptop}  & \textbf{TV}     \\ \hline
pBLEU &0.08 & -0.12 &   0.05\\ \hline
pBBLEU & 0.52 & 0.32  &  0.45   \\ \hline
pBLEURT & 0.38 & 0.17 &  0.26  \\ \hline
pBERT precision & 0.33 & 0.14 & 0.36  \\ \hline
pBERT recall & 0.03 &  -0.06 &  0.14    \\ \hline
pBERT F1 & 0.20 & 0.04 & 0.26  \\ \hline
\end{tabular}
%\end{wrapfigure}
%\vspace{-.2in}
\caption{\label{appendix-tab-correlations} Pearson correlation between SACC and common semantic preservation measures when applied to pseudo-references. All correlations
are statistically significant at p $<$ 0.001 .}
\vspace{-0.1in}
\end{table}

We estimate the goodness of the  pseudo versions of  BLEU, Beyond-BLEU, BERT and BLEURT  by examining their correlations with the domain-specific SACC scores on a sample of model outputs from our experiments, as shown in Table~\ref{appendix-tab-correlations}. The correlations show that the pseudo version of Beyond-BLEU~\cite{wieting-etal-2019-beyond} -- pBBLEU -- performs the best across all three domains. Interestingly, pBLEU, despite BLEU's popularity, performs the worst.

\subsection{Before \& After Ranking}
\label{appendix-before-after}

Our results show that ranking by any ranking function significantly and greatly improves performance, with the greatest performance improvements arising from the RF2$_{DA}$ ranking function
for all three domains. 
We calculate \textit{Before Ranking} by averaging all metrics over the entire set of test outputs (test set size X 10 outputs into ranking). When taking averages across all experiments (per, fine-tuned, and specific), average SACC and DAC are significantly higher after ranking. 

\begin{table}[b]
\small
\begin{tabular}{|c|c|c|c|c|c|c|}
    \hline
    \multicolumn{1}{|c|}{\textbf{N}} & \multicolumn{2}{c|}{\textbf{SACC}} & \multicolumn{2}{c|}{\textbf{Perf}} & \multicolumn{2}{c|}{\textbf{DAC}} \\\hline
    & \bf Before & \bf After & \bf Before & \bf After & \bf Before & \bf After \\\hline
            5 & 65.57&85.71 & 9.10&38.88 & 21.10& 54.44   \\
            25 & 76.01&92.19  & 16.39&62.22 & 31.10&79.72\\
            50 & 86.70&96.43 & 29.10&71.94  & 42.00&79.44\\
            100 & 88.71&97.74 & 40&78.61 & 57.00 &80.56\\\hline
    \end{tabular}
    \caption{Few-shot fine-tuning performance with increasing training examples per DA - before and after ranking. DAC = DA accuracy.}
    \label{tab:ftvsnumprompt}
\end{table}

Table ~\ref{tab:ftvsnumprompt} provides more detail on how the ranking affects
the results for few-shot fine-tuning. Comparing Row 1 to Row 4 shows that ranking improves the performance of SACC for 5-shot fine-tuning (85.71) to perform almost as well as 100-shot fine-tuning before ranking (88.71). Ranking also improves the performance of DAC for 100-shot fine-tuning from 57\% to 80.56\%, a huge improvement.

\begin{figure*}[ht!b]
\begin{center}
\includegraphics[width=5in]{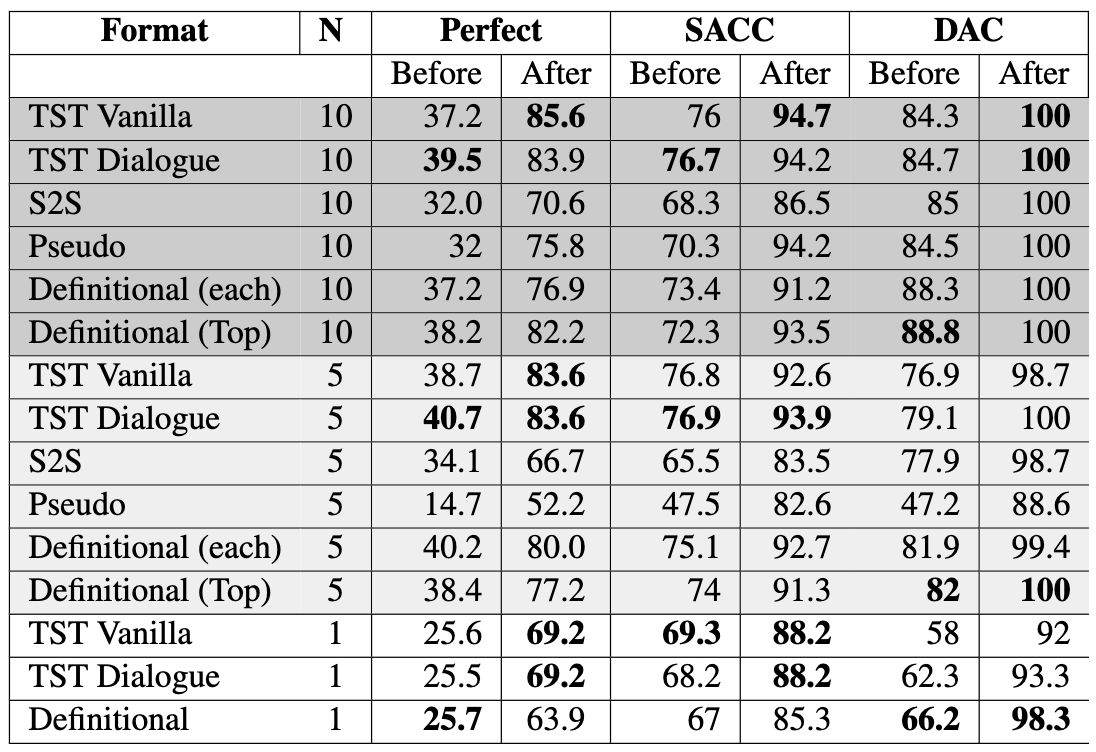}
\end{center}
\caption{Results Before and After Ranking}
\label{tab:before-after-specific_performance}
\end{figure*}

\begin{table*}[h!tb]
\centering
\scalebox{0.8}{\centering\resizebox{\textwidth}{!}{
%\begin{tabular}{ |c|{0.2\linewidth}  c|c{0.2\linewidth} | cc{0.2\linewidth} | rr{0.2\linewidth} |}
\begin{tabular}{ |c|c|c|c|c|c|c|c|}
    \hline
    \multicolumn{1}{|c|}{\textbf{Format}} & 
    \multicolumn{1}{|c|}{\textbf{N}} & \multicolumn{2}{c|}{\textbf{SACC}} & \multicolumn{2}{c|}{\textbf{Perf}} & \multicolumn{2}{c|}{\textbf{DAC}} \\\hline & 
    & \bf Before & \bf After & \bf Before & \bf After & \bf Before & \bf After \\\hline
            TV & 10 & 92.59 & 99.76 & 65.30 &98.85 & 95.90 & 100\\ 
             Laptop & 10 & 80.73 & 95.90 & 36.35 & 80.95 & 99.71 &  100 \\
            \hline
    \end{tabular}}}
    \caption{Laptop and TV Before and After ranking. DAC = DA Accuracy.}
    \label{tab:laptoptvbefore}
\end{table*}

Figure ~\ref{tab:before-after-specific_performance} shows more detail for Viggo across all the experimental settings. \textit{Before Ranking} has an average of 65.29\% versus \textit{After Ranking} with an average of 86.82\% for SACC. DAC has an almost a 30\% increase where \textit{Before Ranking} has an average of 62.11\%, and \textit{After Ranking} has an average of 91.04\%. 
%Diversity metrics are given in lexicalized (Vlex) and delexicalized (Vdlex) vocabulary counts. The original ViGGO training data vocabulary is 1292 lex and 951 dlex. 
Table~\ref{tab:laptoptvbefore} shows the effect of ranking for TV and Laptop, illustrating a similarly large performance improvement due to ranking.

\end{document}